\pdfoutput=1

\documentclass[11pt]{article}

\usepackage[preprint]{acl}

\usepackage{times}
\usepackage{latexsym}
\usepackage{booktabs}
\usepackage{ulem}
\usepackage{enumitem}
\usepackage{soul} 
\usepackage{color, xcolor}
\definecolor{myred}{RGB}{254, 244, 245}
\definecolor{myblue}{RGB}{244, 246, 252}

\usepackage{tcolorbox}
\usepackage{multirow}
\usepackage{amsmath}
\usepackage{amsfonts}  
\usepackage{amssymb}  
\usepackage{caption} 

\usepackage[T1]{fontenc}

\usepackage[utf8]{inputenc}

\usepackage{microtype}

\usepackage{inconsolata}

\usepackage{graphicx}
\usepackage{CJKutf8}

\title{Control-R: Towards controllable test-time scaling}

\author{
 \textbf{Di Zhang\textsuperscript{1, 2}},
 \textbf{Weida Wang\textsuperscript{3, 2}},
 \textbf{Junxian Li\textsuperscript{4}},
 \textbf{Xunzhi Wang\textsuperscript{5}},
 \textbf{Jiatong Li\textsuperscript{6}},
\\
  \textbf{Jianbo Wu\textsuperscript{7}},
 \textbf{Jingdi Lei\textsuperscript{2}},
 \textbf{Haonan He\textsuperscript{8}},
 \textbf{Peng Ye\textsuperscript{2}},
 \textbf{Shufei Zhang\textsuperscript{2}},
\\
 \textbf{Wanli Ouyang\textsuperscript{2}},
 \textbf{Yuqiang Li\textsuperscript{2}},
 \textbf{Dongzhan Zhou\textsuperscript{2}}
\\
 \textsuperscript{1}Fudan University,
 \textsuperscript{2}Shanghai Artificial Intelligence Laboratory,
  \textsuperscript{3}Tongji University,
\\
 \textsuperscript{4}Shanghai Jiaotong University,
 \textsuperscript{5}Nankai University,
  \textsuperscript{6}Hong Kong Polytechnic University,
\\
 \textsuperscript{7}University of California, Merced,
 \textsuperscript{8}University of Science and Technology of China,
\\
 \small{
   \textbf{Correspondence:} \href{mailto:di.zhang@ustc.edu}{di.zhang@ustc.edu}
 }
}

\begin{document}
\maketitle
\begin{abstract}
This paper target in addressing the challenges of underthinking and overthinking in long chain-of-thought (CoT) reasoning for Large Reasoning Models (LRMs) by introducing Reasoning Control Fields (RCF)—a novel test-time approach that injects structured control signals to guide reasoning from a tree search perspective. RCF enables models to adjust reasoning effort according to given control conditions when solving complex tasks. 
Additionally, we present the Control-R-4K dataset, which consists of challenging problems annotated with detailed reasoning processes and corresponding control fields. To further enhance reasoning control, we propose a Conditional Distillation Finetuning (CDF) method, which trains models—particularly Control-R-32B—to effectively adjust reasoning effort during test time. 
Experimental results on benchmarks such as AIME2024 and MATH500 demonstrate that our approach achieves state-of-the-art performance at the 32B scale while enabling a controllable Long CoT reasoning process (L-CoT). Overall, this work introduces an effective paradigm for controllable test-time scaling reasoning.
\end{abstract}
\section{Introduction}

\begin{figure}
\includegraphics[width=0.5\textwidth,height=\textheight,keepaspectratio]{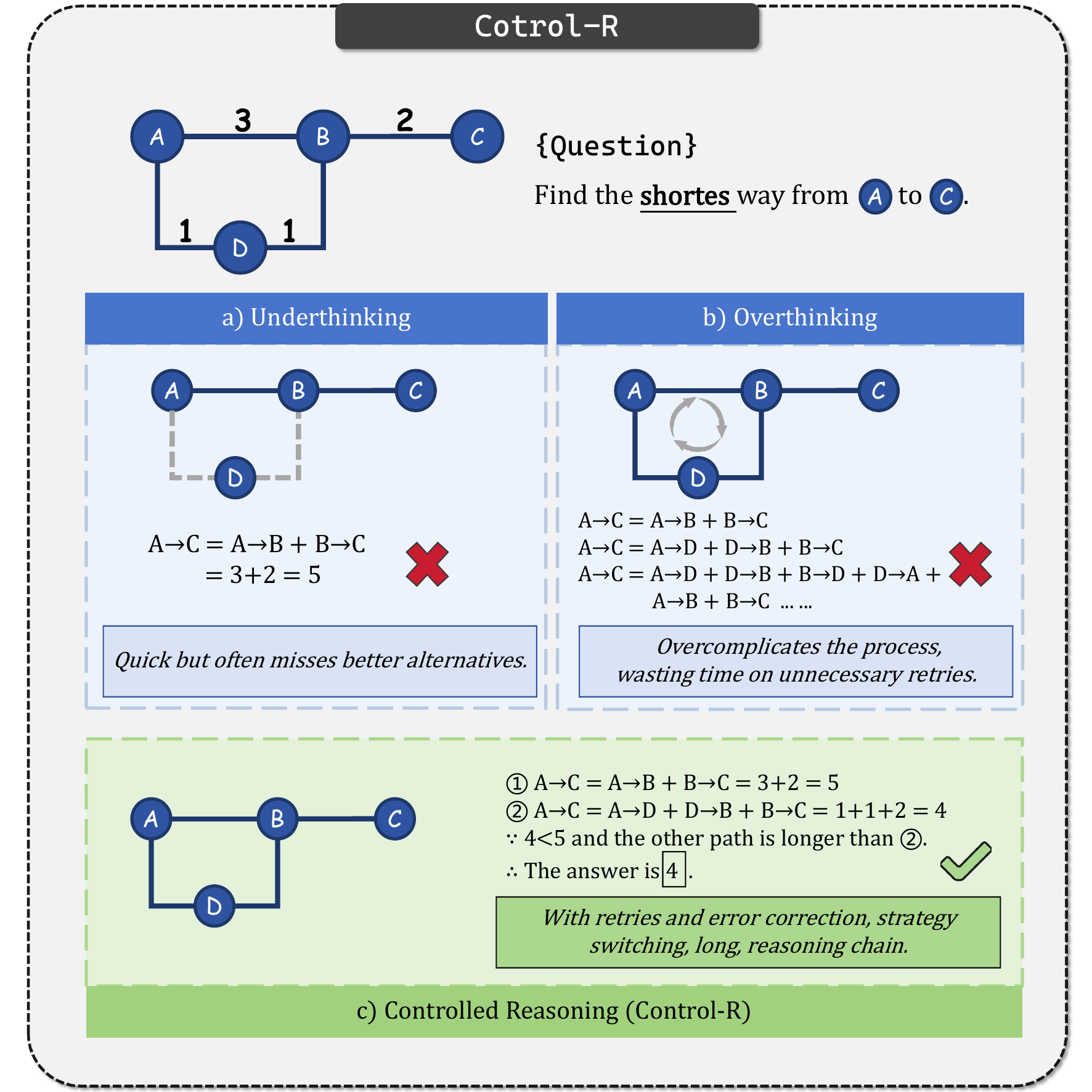} 
\caption{The main idea of Control-R.}
\label{fig:First}
\end{figure}

\textit{“The two methods that seem to scale arbitrarily in this way are search and learning.”}

\hfill — \textit{Rich Sutton, The Bitter Lesson}

Recent advancements in large reasoning models (LRMs), such as DeepSeek-R1~\cite{deepseekai2025deepseekr1incentivizingreasoningcapability}, QwQ-32B~\cite{zheng2024processbenchidentifyingprocesserrors}, and the OpenAI O1 Series~\cite{jaech2024openai}, have enabled 
LRMs to perform reasoning on par with human experts in fields like mathematics and programming. 
However, existing LRMs still face the challenge of handling long chain-of-thought~(L-CoT) reasoning. On one hand, LRMs might fail to fully explore all the possible solutions due to the inherent complexity of the problem, namely \textbf{underthinking}. 
On the other hand, there is still a chance that LRMs engage and stuck in redundant or excessive reasoning, also known as \textbf{overthinking}~\cite{kumar2025overthinking,wang2025thoughts,anderson2025phd}.

As shown in Figure~\ref{fig:First}, in the scenario of \textbf{underthinking}, LRMs tend to rush to a conclusion, without exploring all potential avenues.
This shallow reasoning process, characterized by a lack of self-questioning and reflection, may lead the model to overlook potential errors. 
Conversely, when the model over-elaborates on its reasoning, repeating arguments in an attempt to cover every possible scenario, it falls into \textbf{overthinking}. 
Such overthinking not only consumes excessive computational resources but also reduces the overall efficiency and the quality of the final decision.

Inspired by the ideas of controllable conditional generation behind SteerLM~\cite{dong2023steerlm}, we propose the Reasoning Control Fields~(RCFs) technique as a novel feasible solution. The core concept of RCF is to inject structured control fields into the model during the test-time, thereby enabling efficiency control of the chain-of-thought depth and efficiency. By considering L-CoT reasoning as analogous to a search tree, we decompose both \textbf{execution control} and the \textbf{quality evaluation} of the reasoning process into several attributes, resulting in the definition of 11 distinct RCFs. Each RCF acts as a switch or regulator that guides the model on the extent to which it should engage in strategies such as exhaustive search~\cite{wang2024large}, backtracking~\cite{singhimproving}, or self-refine during its reasoning process~\cite{zhang2024accessing}.

To enable the model to effectively utilize these RCFs, we have constructed a novel, high-quality dataset for L-CoT reasoning, Control-R-4K. This dataset is specifically designed to enhance the model’s search capabilities, providing a range of challenging problems such as\textit{ 24-point game scenarios}, \textit{advanced calculus calculations}, and \textit{differential equation solving}.
Moreover, the dataset is meticulously annotated to ensure clarity and consistency. Each problem is accompanied by detailed reasoning process annotations as well as corresponding control field specifications, making it a comprehensive resource for developing and refining the model’s reasoning abilities.
During training, the model learns to distinguish when deep reasoning is necessary and when redundant steps can be omitted. 

To further enhance this adaptability, we introduce Conditional Distillation Finetuning~(CDF), where the model is trained with both task content and RCFs that guides its reasoning process from tree search perspective. This enables the model to effectively adjust its chain-of-thought depth and strategy based on specified conditions. during test time, users can define control field parameters, allowing the model to adjust reasoning depth and efficiency in test-time.

Experimental results on benchmarks like AIME2024 ~\cite{patel2024aime} and MATH500 ~\cite{lightman2023lets} show that Control-R-32B with RCFs achieve state-of-the-art (SOTA) performance by enhancing search efficiency and reasoning quality. RCFs modulates reasoning effort during test time, which allows users to balance reasoning depth and efficiency dynamically, enabling models to engage in deeper reasoning, reflection, and evidence gathering for complex tasks while streamlining reasoning in simpler cases by reducing redundant steps. 

Overall, our main contributions are as follows:

\begin{itemize}[itemsep=2pt,topsep=0pt,parsep=0pt]
    \item[1.] We propose Reasoning Control Field (RCF) method, which leverages a structured control fields at the test-time of LRMs to efficiently control the long chain-of-thought reasoning. 
    \item[2.] We also introduce the Control-R-4K dataset, featuring questions that require Long chain-of-thought reasoning along with curated reasoning control field annotations to support effective model distillation. 
    \item[3.] Finally, we propose and leverage Conditional Distillation Finetuning~(CDF) for the training of LRMs. Experimental results show that our trained Control-R-32B demonstrated a competitive reasoning ability through controllable and interpretable Long Chain-of-Thought resoning process.
\end{itemize}
\section{Method}

\begin{figure*}[t] 
\centering
\includegraphics[width=\textwidth,height=\textheight,keepaspectratio]{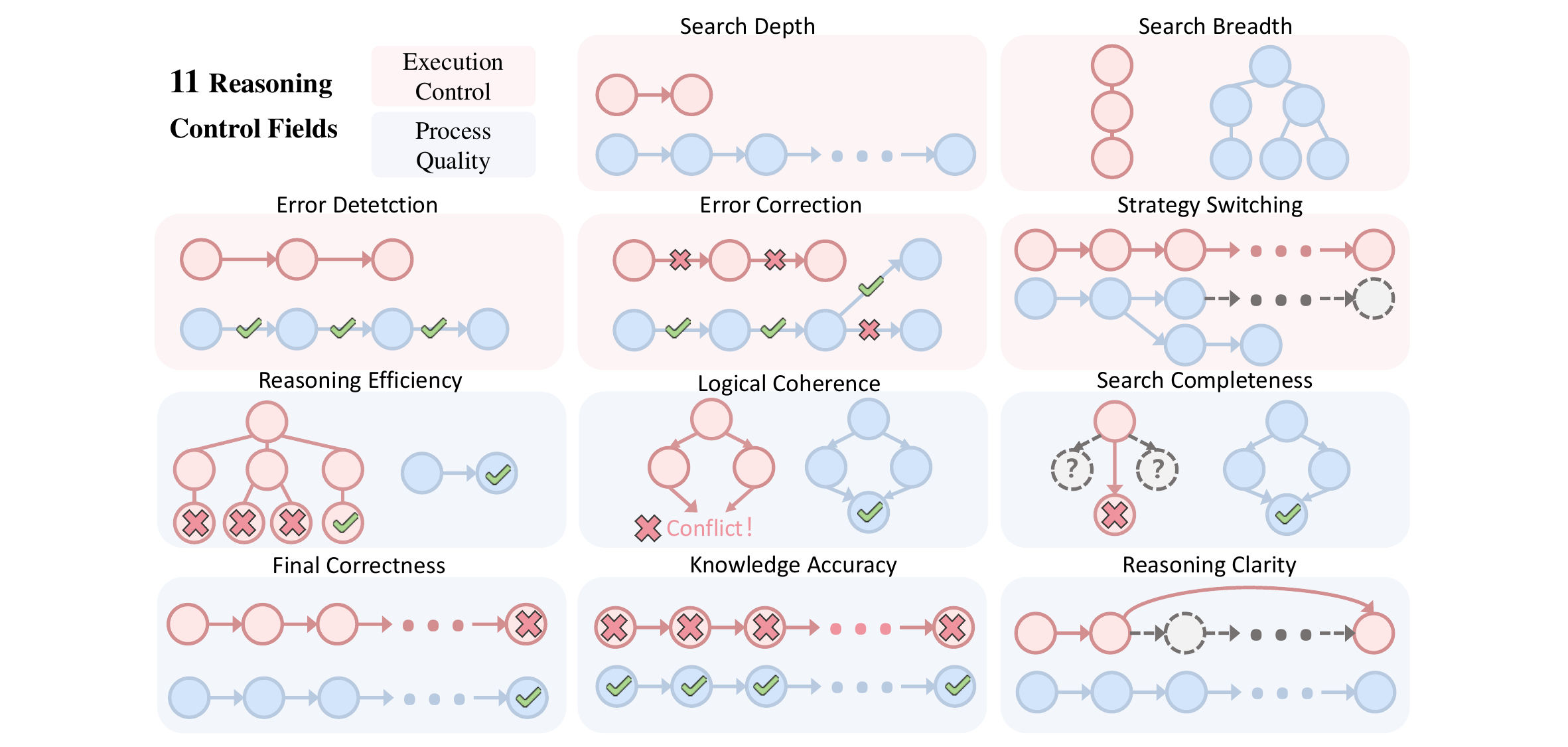} 
\caption{For the explanation of 11 {Reasoning Control Fields}. The \textcolor{red}{red} {L-CoT} represents the tree of thought with low score in this metric, and the \textcolor{blue}{blue} {L-CoT} indicates the high score. The \colorbox{myred}{red} box represents {Execution Control fields}, while the \colorbox{myblue}{blue} box represents {Process Quality Metrics fields}.
}
\label{fig:11RCF}
\end{figure*}

\subsection{Attributes from a Perspective of Tree Search Process}
To control the reasoning effort of LRMs, we propose the RFC technique, which is inspired by the idea behind SteerLM~\cite{dong2023steerlm}. Specifically, our core idea is based on treating the Long CoT~\cite{jaech2024openai} reasoning process of models as an In-Context Tree Search~(ICTS)~\cite{rubin2021learning} process. From this perspective, the L-CoT reasoning process of large reasoning models can be interpreted as a search tree \( \tau(q) \) for a given input query \( q \), where the large reasoning model generates a L-CoT response \( R = (r_1, r_2, \dots, r_n) \). This reasoning process can be formalized as a search tree \( \tau(q) \) with the following properties:

\begin{itemize}[itemsep=2pt,topsep=0pt,parsep=0pt]
    \item Each reasoning step \( r_i \) corresponds to a node \( N_i \) in the search tree \( \tau(q) \).
    \item Different reasoning strategies, such as divide-and-conquer, strategy switching, and error correction methods, influence the selection of exploration paths and nodes to expanse.
    \item Error detection and correction are analogous to backtracking and node selection in tree search process, ensuring the quality and completeness of the L-CoT response $R$.
\end{itemize}

Then we define a set of structured RCFs based on the topology and time-varying characteristics and qualitative metrics of the search tree \( \tau(q) \). As shown in Figure~\ref{fig:11RCF}, these reasoning control fields are categorized into two broad types: \textbf{Execution control metrics} describe how the model control the execution flow of search process. \textbf{Process quality metrics} assess the quality of the steps and connections in search process. Both types of metrics adopt integer scores ranging from 0 to 9, with 0 being the lowest and 9 the highest.

\subsubsection{Execution Control Metrics}

\noindent\textbf{Search depth} gauges the extent of exploration of reasoning. A shallow search (score of 0) signifies minimal, superficial steps, while a deep search (score of 9) involves multi-layered reasoning—employing techniques like recursion and divide-and-conquer—to unravel complex causal relationships.

\noindent\textbf{Search breadth} measures the ability to explore multiple possibilities simultaneously. A narrow approach (score of 0) follows a single and linear path, whereas a broad search (score of 9) systematically compares various solutions using methods such as classification, discussion, and dialectical reasoning.

\noindent\textbf{Error detection} reflects the sensitivity to identify mistakes. A low score (0) indicates a tendency to overlook errors or misclassify correct information as faulty, while a high score (9) is achieved through rigorous self-checks and accurate and frequent identification of errors.

\noindent\textbf{Error correction} evaluates how effectively the system rectifies identified mistakes. Minimal corrective action (score of 0) can lead to compounding errors, whereas a high level of error correction (score of 9) is marked by active backtracking and iterative adjustments that ensure precision.

\noindent\textbf{Strategy switching} assesses the flexibility in adapting the search process. Ineffective strategy switching (score of 0) may result in erratic changes or an inflexible adherence to one method, while optimal strategy switching (score of 9) involves targeted shifts between various approaches to enhance search efficiency and overall solution quality.
  
\subsubsection{Process Quality Metrics}

\noindent\textbf{Final correctness} measures how accurately the conclusion aligns with established facts or the standard solution. A score of 0 indicates major errors and a severe deviation from accepted answers, while a score of 9 reflects complete accuracy and high reliability.

\noindent\textbf{Reasoning efficiency} evaluates the use of resources and steps during the problem-solving process. A score of 0 suggests excessive, redundant work and inefficient use of resources, whereas a score of 9 signifies that high-quality results were achieved with minimal steps and optimal focus.

\noindent\textbf{Search completeness} examines whether all critical conditions, arguments, and steps are addressed. A score of 0 denotes significant omissions, while a score of 9 confirms that all essential elements were thoroughly considered.

\noindent\textbf{Logical coherence} reviews the consistency and rigor of the reasoning process. A score of 0 is given when the reasoning is contradictory or confused, whereas a score of 9 indicates a tightly connected, well-structured, and internally consistent argument.

\noindent\textbf{Knowledge accuracy} focuses on the correctness of the facts, concepts, or theories referenced. A score of 0 points to multiple errors or heavy hallucinations, while a score of 9 demonstrates a precise and solid grasp of the relevant knowledge or facts.

\noindent\textbf{Reasoning clarity} assesses how clearly the reasoning is presented. A score of 0 indicates a chaotic or confusing explanation with missing steps, and a score of 9 reflects a logically sound, well-organized presentation that is easy to follow.

\subsection{Reasoning Control Fields}

Inspired by SteerLM, we adopt the concept of conditional generation in probability theory by introducing attribute control information at each token generation step. This approach ensures that the generated content is not merely a response to the input prompt but also adheres to predefined structured attribute fields requirements, forming high-quality reasoning process. To achieve this, we introduce a structured reasoning control fields. Given a reasoning task \( q \) with a corresponding reasoning process \( R \), we represent the attribute requirements during reasoning or search as a structure of reasoning control fields \( C \), which comprises eleven predefined metrics that define attribute information throughout the L-CoT reasoning process. 

From a probabilistic perspective, without additional constraints, the model generates the reasoning process \( R = (r_1, r_2, \ldots) \) step by step freely, where each token depends solely on the previously generated tokens. The probability of generating the entire reasoning process can be expressed as:

\begin{small}
\begin{equation}
P(R\mid q)=\prod_{t=1}^{T}P(r_t\mid r_1, r_2, \ldots, r_{t-1}, q).
\end{equation}
\end{small}

When the reasoning control fields \( C \) is introduced, we require that the generated reasoning process not only matches the task \( Q \) but also satisfies the attribute requirements defined in reasoning control fields \( C \). Thus, the generation process becomes conditional on both \( Q \) and \( C \), and the probability of model's generation process is modified as follows:

\begin{small}
\begin{equation}
P(R\mid q, C)=\prod_{t=1}^{T}P(r_t\mid r_1, r_2, \ldots, r_{t-1}, q, C).
\label{Eq:RQC}
\end{equation}
\end{small}

This equation indicates that each token \( r_t \) is generated based on the previously generated tokens, the query of task \( q \), and the attributes specified in the control fields \( C \), thereby enabling a control on the model's long chain-of-thoutht reasoning process.

\begin{figure*}[t] 
\centering
\includegraphics[width=\textwidth,height=\textheight,keepaspectratio]{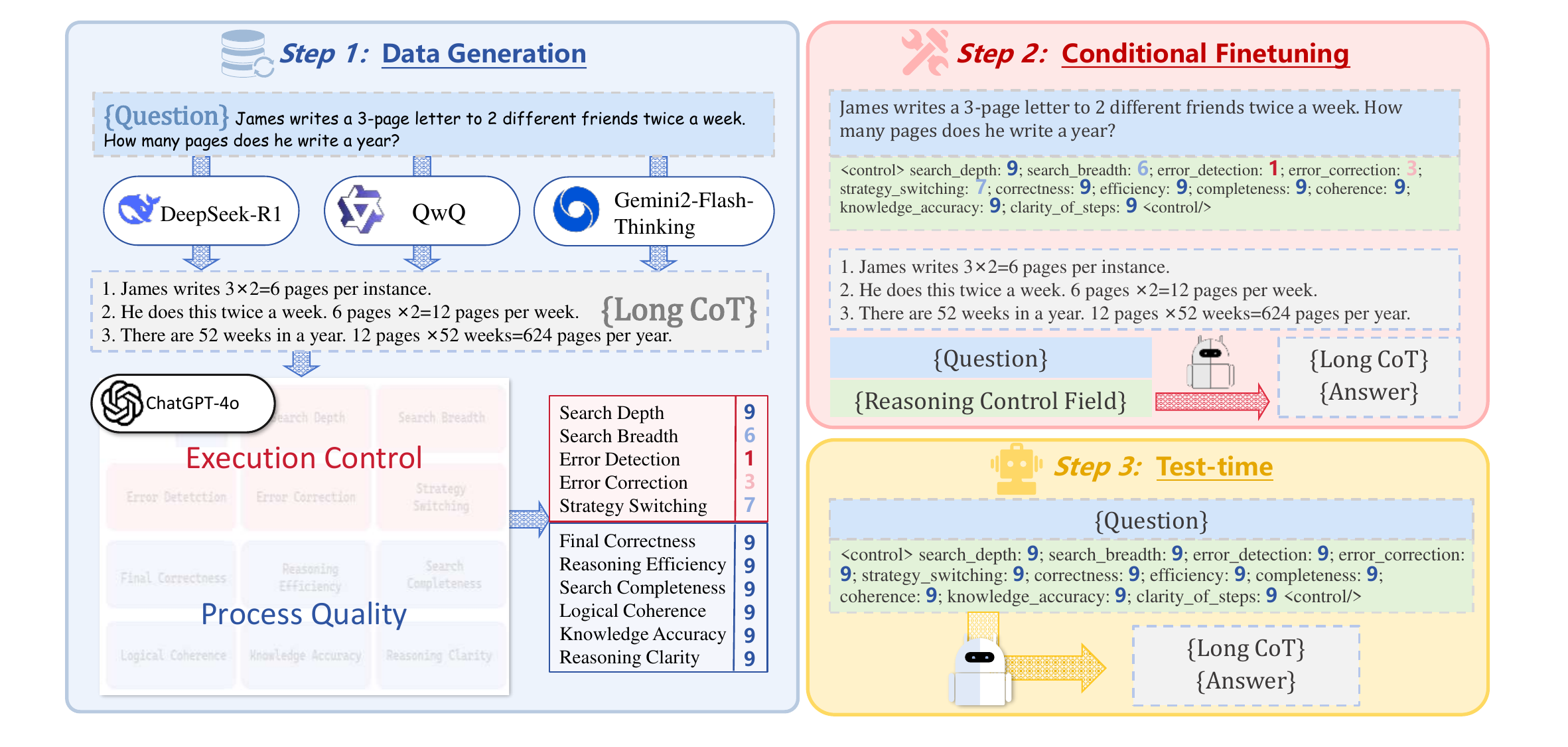} 
\caption{The overall pipeline of Control-R.}
\label{fig:pipeline}
\end{figure*}

\textbf{Attribute Annotation: }To systematically describe the Reasoning Control Fields (RCFs) in long Chain-of-Thought (L-CoT) reasoning processes, we introduce a structured attribute annotation format. This structured representation allows for detailed evaluation and assessment over different aspects of reasoning. The annotation is defined as following example:  

\begin{small}
\begin{quote}
\begin{verbatim}
{
  "analysis": {
    "execution_control_scores": {
      "search_depth": 8,
      "search_breadth": 7,
      "error_detection": 8,
      "error_correction": 7,
      "strategy_switching": 6
    },
    "quality_evaluation_scores": {
      "correctness": 9,
      "efficiency": 7,
      "completeness": 8,
      "coherence": 8,
      "knowledge_accuracy": 9,
      "clarity_of_steps": 8
    },
    "justification": "The reasoning..."
  }
}
\end{verbatim}
\end{quote}
\end{small}

As illustrated in Figure~\ref{fig:pipeline}, to obtain reasoning attribute annotation in this format, we prompted ChatGPT-4o to annotate Long Chain-of-Thoughts reasoning processes generated by state-of-the-art large reasoning models, including DeepSeek-R1, QwQ-32B, and Gemini-2-Flash-Thinking. These structured annotations form serve as the control signal in our proposed distillation training method, ensuring models can learn to control of reasoning behaviors accordingly.  

To integrate these control signals into model training, we thus convert the structured attribute annotations into a textual format, referred to as reasoning control fields~(RCFs) strings. These fields are then appended to the user query as a guiding constraint for reasoning process in the training dataset:  

\begin{small}
\begin{quote}
\begin{verbatim}
"\n<control> search_depth: 8; search_breadth:
7; error_detection: 8; error_correction: 7; 
strategy_switching: 6; correctness: 9; 
efficiency: 7; completeness: 8; coherence: 8; 
knowledge_accuracy: 9; clarity_of_steps: 8 
<control/>"
\end{verbatim}
\end{quote}
\end{small}

During training time, the model learns to incorporate these RCFs  strings as constraints as discussed in~\eqref{Eq:RQC}, ensuring that the generated reasoning processes align with both the query requirements and the predefined reasoning attributes. For details of the prompting strategies used to obtain these annotations, refer to Appendix~\ref{sec:Details of Prompt of Attribute annotations}. Additionally, the dataset employed in the annotation process, Control-R-4K, is described in Appendix~\ref{sec:Details of Dataset Control-R-4K}. 

\subsection{Conditional Distillation Fine-tuning}

To effectively teach the model to reason in the Long Chain-of-Thought style while adhering to the specific conditions of appended RCFs string, we applied Conditional Distillation Fine-tuning~(CDF) using our newly curated Control-R-4K dataset. Given a query \( q \), the model generates a reasoning process \( R \), which consists of a sequence of reasoning steps:

\begin{equation}
R = \{ r_1, r_2, \dots, r_n \}, \quad r_i \in \mathcal{R},
\end{equation}

\noindent where \( \mathcal{R} \) is the space of all possible reasoning steps. Each reasoning process $R$ can be annotated with an reasoning control field \( C \), a textual representation of different reasoning attributes:

\begin{equation}
C = (c_1, c_2, \dots, c_k) \in \mathcal{C}^k,
\end{equation}

\noindent where \( k \) is the number of distinct control attributes. Thus, a fully annotated reasoning process can be represented as $(q, R, C)$.

A key challenge in directly training the model on multiple annotated processes is the potential for conflicts between different reasoning strategies for the same query. Formally, given a dataset \( \mathcal{D} \) consisting of multiple valid reasoning processes for the same query:

\begin{small}
\begin{equation}
\mathcal{D} = \{ (q, R_1, C_1), (q, R_2, C_2), \dots, (q, R_m, C_m) \},
\end{equation}
\end{small}

\noindent where $q$ is fixed, but \( R_i \neq R_j \) for \( i \neq j \), the model may struggle to reconcile different reasoning styles described by $C_i \neq C_j$, leading to inconsistent behavior when no explicit control field is provided during inference.

To address this, we introduce Conditional Distillation Fine-tuning (CDF), which conditions the model on attribute control fields \( C \) to ensure that it learns controllable reasoning processes rather than arbitrarily averaging multiple reasoning styles. The training objective for CDF is formulated as:

\begin{small}
\begin{equation}
\mathcal{L}_{\text{CDF}} = \mathbb{E} \left[ -\sum_{i=1}^{|R|} \log P(r_i \mid r_{<i}, q, C; \theta) \right]
\end{equation},
\end{small}

\noindent where \( P(r_i \mid r_{<i}, q, C; \theta) \) is the probability assigned by the model which parameterized by \( \theta \) to the next reasoning step \( r_i \), given the previous steps and the control field \( C \). This ensures that the model explicitly conditions on \( C \) when generating reasoning processes, leading to a controllable and interpretable L-CoT reasoning process.



\section{Results}
\begin{figure}[t] 
\centering
\includegraphics[width=0.45\textwidth,height=\textheight,keepaspectratio]{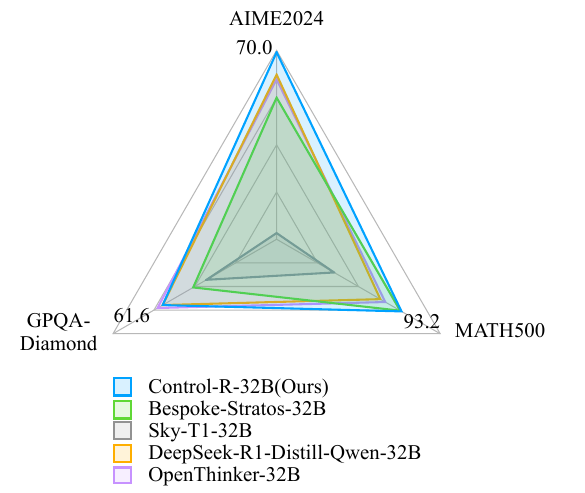} 
\caption{Radar chart of experiments results on cutting-edge competition benchmarks.}
\label{fig:pdf_insert}
\end{figure}
\subsection{Evaluation Settings}
We fine-tuned the Qwen2.5-32B-Instruct model using the Control-R-4K dataset, resulting in the Control-R-32B model. For hyper-parameter setting in training, please refer to Appendix~\ref{sec:Hyper-parameter Settings in training}. Subsequently, we evaluated this model based on the following prompting strategy:

\noindent \textbf{Generation: }The system prompt is set to an empty string, while the user prompt consists of the original question, followed by control fields and an inference prompt as shown below:

\begin{center}
\begin{small}
\begin{quote}
\begin{verbatim}
\n<control> search_depth: 9; search_breadth: 9;
error_detection: 9; error_correction: 9; 
strategy_switching: 9; correctness: 9; 
efficiency: 9; completeness: 9; 
coherence: 9; knowledge_accuracy: 9; 
clarity_of_steps: 9 <control/>
Please reason step by step, and put your final 
answer within \\boxed{}.
\end{verbatim}
\end{quote}
\end{small}
\end{center}

Notably, following the suggestion of DeepSeek research team~\cite{guo2025deepseek}, we incorporate the following forced generation prompt to avoid the model’s tendency to skip reasoning process:

\noindent \begin{small}
\begin{quote}
\begin{verbatim}
<think>\n
\end{verbatim}
\end{quote}
\end{small}

\noindent \textbf{Grading: }We first extract the predicted answers from the model's responses using regular expressions. Then, we employ the grading tool from PRM800K\cite{lightman2023lets} to assess the equivalence between the predicted and ground truth answers. If extraction fails or the answers are deemed non-equivalent, the response is marked as incorrect.

\noindent \textbf{Metric: }All performance metrics are computed using the \texttt{Pass@1} metric~\cite{chen2021evaluating}, which is defined as:  
\[
\mathrm{pass@}k:=\underset{\mathrm{Problems}}{\operatorname*{\operatorname*{\mathbb{E}}}}\left[1-\frac{\binom{n-c}{k}}{\binom{n}{k}}\right]
\]
where $n$ is the number of sampled responses, and $c$ is the number of correct responses.

\subsection{Main Result}

\begin{table*}[!t]
\centering
\resizebox{0.75\linewidth}{!}{
\begin{tabular}{lcccc}
\toprule
\multirow{2.5}{*}{\textbf{Model}} & \multicolumn{3}{c}{\textbf{Cutting-edge}} & \multicolumn{1}{c}{\textbf{Uncontaminated}} \\
\cmidrule(lr){2-4} \cmidrule(lr){5-5}
               & {AIME2024} & {MATH500} & {GPQA-Diamond} & {AIME2025 Part I} \\
\midrule

{Bespoke-Stratos-32B}               & 63.3  & 93.0  & 58.1  & -  \\
{Sky-T1-32B}                        & 43.3  & 82.4  & 56.8  & -  \\
{DeepSeek-R1-Distill-Qwen-32B}      & 66.7  & 89.8  & 61.1  & 53.3  \\
{OpenThinker-32B}                   & 66.0  & 90.6  & 61.6  & 53.3  \\
\midrule
{Control-R-32B(Ours)}         & \underline{70.0}  & \underline{93.2}  & 61.1  & 55.0  \\
\midrule
{o1-preview}                        & 40.0  & 81.4  & \textbf{75.2}  & \textbf{78.3}  \\
{DeepSeek-R1}                       & \textbf{79.8}  & \textbf{97.3}  & \underline{71.5}  & \underline{65.0}  \\

\bottomrule
\end{tabular}
}
\caption{Experiments on cutting-edge benchmarks, including AIME2024, MATH500, GPQA-Diamond, and AIME2025 Part I. The metric is Pass@1 (n=1). Reported results cited from Bespoke-Stratos Project~\cite{bespoke_stratos}.}
\label{tab:main_table}
\end{table*}

\textbf{On cutting-edge competition benchmarks. }As shown in Table~\ref{tab:main_table}, Control-R-32B (Ours) demonstrates a strong performance across benchmarks, excelling in mathematical reasoning. 

In AIME2024, it achieves 70.0\%, outperforming Bespoke-Stratos-32B (63.3\%) and Sky-T1-32B (43.3\%) but trailing DeepSeek-R1 (79.8\%). In MATH500, it leads with 93.2\%, surpassing OpenThinker-32B (90.6\%) and Bespoke-Stratos-32B (93.0\%), while DeepSeek-R1 (97.3\%) scores highest. And in GPQA-Diamond, it scores 61.1\%, outperforming Bespoke-Stratos-32B (58.1\%) and Sky-T1-32B (56.8\%), but lags behind models with larger parameter sizes such as o1-preview (75.2\%) and DeepSeek-R1 (71.5\%).  

In conclusion, Control-R-32B achieves SOTA performance on MATH500 and AIME2024 at 32B scale while remaining competitive in other tasks. It demonstrates stable performance, whereas some models excel in specific benchmarks but underperform elsewhere. Future work should enhance its performance for improved complex reasoning in scientific areas.

\noindent \textbf{On latest uncontaminated competition benchmarks. }To analyze the performance of the model on the latest evaluation benchmarks that are least contaminated by pretraining data, we conducted tests and comparisons on AIME2025 Part I. As shown in Table~\ref{tab:aime2025}, Control-R-32B (Ours) achieved an average accuracy of 55\% in the AIME2025 Part I benchmark. While it demonstrated strong performance in certain tasks, there remains significant room for improvement.  

Control-R-32B exhibited performance comparable to the closed-source model Gemini-2.0-Flash-Thinking (51.67\%), though differences were observed across individual tasks. Compared with similar parameter size models such as QwQ-32B (36.67\%), Control-R-32B demonstrated competitive experimental results. Compared to traditional LLMs, Control-R-32B outperformed DeepSeek-V3 (28.33\%), Gemini-2.0-Flash (30\%), and Claude-3.5-Sonnet (3.33\%), demonstrating a certain level of competitiveness. 

In summary, Control-R-32B exhibits a certain level of competitiveness and performs well on similar parameter sizes, but there remains considerable room for improvement. By optimizing training strategies, enhancing reasoning capabilities, and improving generalization in future works, it is expected to narrow the gap with the O3 and O1 series, further enhancing overall performance.

\subsection{Ablation Study}
\begin{table*}[!t]
\centering
\small 
\resizebox{0.75\textwidth}{!}{ 
\begin{tabular}{ccccc ccc}
\toprule
\multirow{2.5}{*}{\textbf{Dataset}} & \multirow{2.5}{*}{\textbf{Outcome}} & \multicolumn{3}{c}{\textbf{Token Length}} & \multicolumn{3}{c}{\textbf{Wait Occurrence Times}} \\ 
\cmidrule(lr){3-5} \cmidrule(lr){6-8}
& & \textbf{Longest} & \textbf{Shortest} & \textbf{Avg} & \textbf{Most} & \textbf{Least} & \textbf{Avg} \\ 
\midrule
\multirow{2}{*}{Math500}  & Correct & 24375 & 611 & 1687.64 & 123 & 0 & 5.64 \\
                          & Wrong   & 21167 & 1232 & 281.59  & 131 & 2 & 1.74 \\ 
\midrule
\multirow{2}{*}{GPQA-Diamond}  & Correct & 21740 & 377 & 806.42 & 226 & 0 & 7.02 \\
                               & Wrong   & 26155 & 558 & 654.13 & 315 & 0 & 6.53 \\ 
\midrule
\multirow{2}{*}{AIME2024}  & Correct & 22980 & 2024 & 185.61 & 160 & 5 & 1.05 \\
                           & Wrong   & 22315 & 6859 & 173.55 & 255 & 32 & 1.60 \\ 
\midrule
\multirow{2}{*}{AIME2025}  & Correct & 9179 & 1637 & 41.05 & 38 & 2 & 0.14 \\
                           & Wrong   & 26585 & 4617 & 154.79 & 250 & 34 & 1.10 \\ 
\bottomrule
\end{tabular}
}
\caption{Summary of Token Length and "Wait" Occurrence for Different Datasets}
\label{tab:dataset_summary}
\end{table*}

\textbf{A Statistical Perspective on the L-CoT Reasoning Process.} Table~\ref{tab:dataset_summary} examines token length and "Wait" occurrences across different datasets to explore their impact on reasoning correctness. The results highlight key trends that can inform model optimization.

Correct answers generally involve longer token sequences, particularly in Math500 and GPQA-Diamond, where correct responses have significantly greater average token lengths (1687.64 vs. 281.59 and 806.42 vs. 654.13, respectively). However, this trend reverses in AIME2025, where incorrect answers are longer (154.79 vs. 41.05), suggesting that excessive token generation in complex tasks may indicate redundant computation or reasoning errors, aligning with prior findings~\cite{muennighoff2025s1simpletesttimescaling}.

"Wait" occurrences also impact correctness. In GPQA-Diamond, correct answers exhibit slightly higher average "Wait" occurrences (7.02 vs. 6.53), suggesting that moderate pauses may improve precision. However, in AIME2025, correct answers have minimal "Wait" occurrences (0.14 vs. 1.1), indicating that fluency is more critical in simpler tasks. Additionally, incorrect reasoning often involves extreme "Wait" occurrences, as seen in GPQA-Diamond and AIME2024, where incorrect responses exhibit substantially higher maximum wait occurrences (315 vs. 226 and 255 vs. 160), reflecting inefficiencies or inference loops.

In summary, longer token sequences and moderate pauses generally enhance accuracy, but their effects depend on task complexity. Future optimizations should tailor reasoning strategies to different problem types, balancing thoroughness and efficiency.

\begin{table}[!t]
\centering
\resizebox{1.0\linewidth}{!}{
\begin{tabular}{lll}
\toprule
\textbf{Model}                                                                                         & \textbf{AIME2024} & \textbf{MATH500} \\
\midrule
Qwen2.5-32B-Instruct                                                                              & 13.3     & 81.6    \\
\midrule
\begin{tabular}[c]{@{}l@{}}Control-R-32B without \\ control fields\end{tabular}                   & 6.67     & 3.2     \\
\midrule
\begin{tabular}[c]{@{}l@{}}Control-R-32B with \\ control fields (all set to 0)\end{tabular}       & 63.3     & 91.4    \\
\midrule
\begin{tabular}[c]{@{}l@{}}Control-R-32B with \\ control fields (all set to 5)\end{tabular}       & 63.3     & 91.8    \\
\midrule
\begin{tabular}[c]{@{}l@{}}Control-R-32B with \\ control fields (all set to 9, Ours)\end{tabular} & \textbf{70.0}       & \textbf{93.2}   \\
\bottomrule
\end{tabular}
}
\caption{Ablation study on the affection of reasoning control fields}
\label{tab:ablation_rfc}
\end{table}

\noindent \textbf{On the impact of reasoning control fields: }As Shown in Table~\ref{tab:ablation_rfc}, we examine the impact of reasoning control fields on the model's L-CoT performance by comparing Qwen2.5-32B-Instruct and Control-R-32B on AIME2024 and MATH500.  

Control-R-32B without control fields performs significantly worse (AIME2024: 6.67\%, MATH500: 3.2\%) than the baseline Qwen2.5-32B-Instruct, highlighting the necessity of control fields. With control fields set to 0 or 5, Control-R-32B achieves 63.3\% on AIME2024 and ~91.6\% on MATH500, showing minimal sensitivity in MATH500 but a notable impact on AIME2024. Setting the fields to 9 (Ours) further improves performance to 70.0\% on AIME2024 and 93.2\% on MATH500, indicating that optimized control configurations enhance reasoning performance. These results suggest that AIME2024 is more sensitive to control field configurations than MATH500. These findings underscore the need for task-specific control strategies to maximize model performance in future studies.
\section{Related Work}

\paragraph{Reasoning with LLMs.} Recent language models have seen great success in solving math, physics or other complex reasoning problems~\cite{brown2020language, touvron2023llama, hoffmann2022training, chiang2023vicuna}. To enhance reasoning ability of LLMs, instead of making the models generate a single answer directly, previous works have designed a step-by-step reasoning framework through specially designed instructions. Chain-of-Thought~\cite{wei2022chain} and others~\cite{kojima2022large, wang2023plan} are some of the representative ones. To further improve the reasoning process, some recent works focus on preference optimization~\cite{lai2024step, pang2024iterative, lahlou2024port}, involving tree-based searching methods~\cite{xie2024monte, zhang2024accessing}, or adding self-evaluation or verification to the reasoning steps~\cite{weng2023large, xie2023self}. Compared to these works, our method adopt a special attributed conditional method for data construction and training. This makes sure controllable test-time scaling. 

\paragraph{Attributed Conditional Training.} In this paper we include an attributed conditional distillation method for training. The exploration of attriubted conditioning is widely used in controllable image generation~\cite{abdal2021styleflow}. Similar idea in supervised fine-tuning in LLMs is firstly introduced by~\cite{dong2023steerlm}, which ensure users to control responses during inference. For this purpose, special datasets~\cite{wang2024helpsteer, wang2024helpsteer2}are proposed. In our paper, we take the idea of attributed conditional training and propose new methods for enhancing controllable reasoning.
\section{Conclusion}
In this paper, we introduced Reasoning Control Fields (RCF), a test-time approach that injects structured control signals to adjust the long chain-of-thought reasoning process of LRMs. Our contributions include the high-quality Control-R-4K dataset and the Conditional Distillation Finetuning (CDF) strategy, and the Control-R-32B model, which can adjust reasoning efforts according to given reasoning control fields. Experimental results on benchmarks like AIME2024 and MATH500 shows that Control-R-32B with RCF not only achieves state-of-the-art performance at the 32B scale, but also generate a controllable and intepretable L-CoT reasoning process. 

\section*{Limitations}

While our proposed approach demonstrates strong performance in structured reasoning tasks, several areas warrant further exploration:  

\textbf{Generalization to Open-Domain Reasoning: }Our method has been validated on mathematical and logical benchmarks, but its effectiveness in open-domain scenarios (e.g., commonsense, scientific, and legal reasoning) remains to be explored.  

\textbf{Fine-Grained Control Calibration: }While control fields offer flexibility, selecting optimal values for different tasks may require manual tuning. Future work could explore adaptive mechanisms for automated calibration.  

\textbf{Evaluation Scope: }Our experiments focus on key reasoning benchmarks, but broader validation on diverse datasets, including real-world and adversarial tasks, would provide a more comprehensive assessment of robustness and applicability. 



\bibliography{custom}

\clearpage
\appendix

\setcounter{table}{0} 
\renewcommand{\thetable}{A\arabic{table}}
\setcounter{figure}{0}
\renewcommand{\thefigure}{A\arabic{figure}}
\setcounter{equation}{0}
\renewcommand{\theequation}{A\arabic{equation}}
\section{Details of Prompt used of Attribute Annotations}
\label{sec:Details of Prompt of Attribute annotations}

\begin{figure*}
    \centering
    \includegraphics[width=0.95\linewidth]{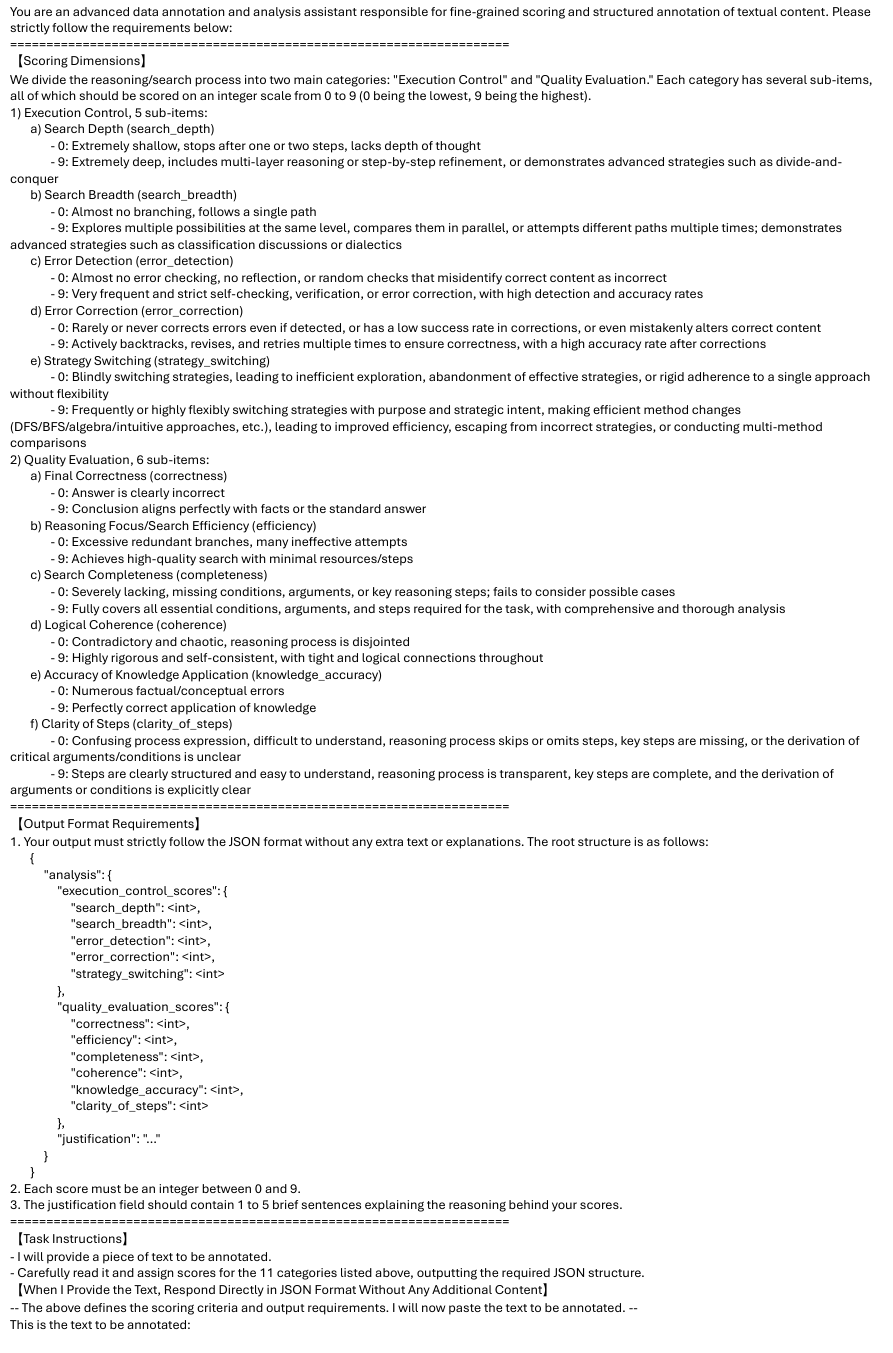}
    \caption{Prompt used of Attribute Annotations}
    \label{fig:prompts}
\end{figure*}

The prompt used in the curation process of Control-R-4K
 to produce attribute annotations was depicted in Figure~\ref{fig:prompts}.
\section{Details of Dataset Control-R-4K}
\label{sec:Details of Dataset Control-R-4K}
\subsection{Dataset Collection}

\begin{table*}[!t]
\resizebox{1.0\linewidth}{!}{
\begin{tabular}{llll}
\toprule
\textbf{subset} & \textbf{capacity} & \textbf{generated from}  &  \textbf{attribute annotation labeler} \\
\midrule
main            & 2610              & Deepseek-R1  & ChatGPT-4o                                \\
search task     & 1895              & Deepseek-R1 &  ChatGPT-4o                               \\
extended        & 35000             & QwQ-32B, Deepseek-R1, Gemini2-Flash-Thinking  & ChatGPT-4o\\
\bottomrule
\end{tabular}
}
\caption{Subsets of Control-R-4K}
\label{tab:Subsets of Control-R-4K}
\end{table*}

As shown in Table~\ref{tab:Subsets of Control-R-4K}, the Control-R-4K dataset consists of three components: main subset, search task subset, and extended subset.

The main subset data comprises AIME exam problems from 1983 to 2023, AMC exam problems from 2022, and a random sample of problems from the MATH dataset.

The search task subset is automatically synthesized using random rules and includes problems on the 24-points Game and calculus calculations. The calculus problems encompass tasks such as differentiation, integration, limit evaluation, and differential equations solving.

The extended subset data is composed of the datasets
\href{https://huggingface.co/datasets/bespokelabs/Bespoke-Stratos-17k}{\texttt{bespokelabs/Bespoke-Stratos-17k}},
\href{https://huggingface.co/datasets/NovaSky-AI/Sky-T1_data_17k}{\texttt{NovaSky-AI/Sky-T1\_data\_17k}}, and
\href{https://huggingface.co/datasets/simplescaling/s1K}{\texttt{simplescaling/s1K}}.

\subsection{Sampling Responses}

The supplementary data already includes responses generated via Long Chain-of-Thought and remains unchanged.

For the primary data and search task data, we perform response sampling using DeepSeek-R1. We maintain the hyperparameters originally recommended for sampling, with an exception of adjusting the sampling temperature to 1.0 to achieve more diverse response data.

The sampling outcomes are evaluated for accuracy using the grading tool from PRM800K~\citep{lightman2023prm800k}, and any incorrect sampling outcomes are discarded.

\subsection{Attribute Annotations}

We utilize ChatGPT-4o to annotate the attributes of the aforementioned response data. The temperature for the annotation process is set to 0.0. The prompt used for the annotation process is as shown in Figure~\ref{fig:prompts}, and a sampled structured representation is as follows,

\begin{small}
\begin{quote}
\begin{verbatim}
{
  "analysis": {
    "execution_control_scores": {
      "search_depth": 8,
      "search_breadth": 7,
      "error_detection": 8,
      "error_correction": 7,
      "strategy_switching": 6
    },
    "quality_evaluation_scores": {
      "correctness": 9,
      "efficiency": 7,
      "completeness": 8,
      "coherence": 8,
      "knowledge_accuracy": 9,
      "clarity_of_steps": 8
    },
    "justification": "The reasoning..."
  }
}
\end{verbatim}
\end{quote}
\end{small}

Then we convert this structure to the control fields as a string in the following pythonic format:

\begin{small}
\begin{quote}
\begin{verbatim}
"\n<control> search_depth: 8; search
_breadth: 7; error_detection: 8; err
or_correction: 7; strategy_switching
: 6; correctness: 9; efficiency: 7; 
completeness: 8; coherence: 8; know
ledge_accuracy: 9; clarity_of_steps
: 8 <control/>"
\end{verbatim}
\end{quote}
\end{small}

\section{Hyper-parameter Settings in training}
\label{sec:Hyper-parameter Settings in training}
The training process employs LoRA (Low-Rank Adaptation)~\citep{hu2021lora}, Flash-Attention-2~\citep{dao2023flashattention}, and Liger Kernel~\citep{hsu2024ligerkernelefficienttriton} for optimization, fine-tuning based on Qwen2ForCausalLM from the Huggingface Transformers library~\footnote{https://github.com/huggingface/transformers/}. The training is distributed across four nodes with a total of 32 A100 GPUs on a slurm cluster (eight GPUs per node) and utilizes the DeepSpeed ZeRO-3~\citep{rajbhandari2020zero} optimization strategy to minimize memory consumption and enhance computational efficiency. The rank of LoRA layers is set to 16, with all linear layers of the model as the targets of LoRA adapters.

For optimization, the training procedure adopts the Adam~\citep{kingma2014adam} optimizer ($\beta_1$ = 0.9, $\beta_2$= 0.999, $\epsilon$ = 1e-8) and BF16 (bfloat16)~\citep{kalamkar2019bf16} precision. The initial learning rate is set to 5e-5, following a cosine learning rate scheduler with a warm-up ratio of 10\%. The total batch size is fixed at 32, and gradient accumulation is not enabled.  

During training, a total of 1,660 steps were completed. At the end of training, the final loss was 0.4431, with an overall average loss of 0.5690, a sample-level loss of 0.9674, and a gradient norm of 0.24096.
\section{Detailed results of AIME2025 Part I}
\label{sec:Detailed results of AIME2025 Part I}
The detailed results of AIME 2025 Part I are shown in Table~\ref{tab:aime2025}.
\begin{table*}[!t]
\resizebox{1.0\linewidth}{!}{
\begin{tabular}{lllllllllllllllll}
\toprule
\textbf{Model}                     & \textbf{Avg. Acc.} & \textbf{1} & \textbf{2} & \textbf{3} & \textbf{4} & \textbf{5} & 6   & 7   & 8   & 9   & 10  & 11  & 12  & 13 & 14 & 15 \\
\midrule
O3-mini (high)                & 80                 & 100        & 100        & 100        & 100        & 75         & 100 & 100 & 100 & 100 & 75  & 100 & 100 & 50 & 0  & 0  \\
\textbf{O1 (medium)}          & 78.33              & 100        & 100        & 100        & 100        & 100        & 100 & 75  & 50  & 100 & 100 & 100 & 100 & 25 & 25 & 0  \\
O3-mini   (medium)            & 73.33              & 100        & 100        & 100        & 100        & 75         & 100 & 50  & 100 & 100 & 100 & 75  & 100 & 0  & 0  & 0  \\
DeepSeek-R1                   & 65                 & 100        & 100        & 100        & 100        & 100        & 100 & 50  & 100 & 75  & 25  & 25  & 100 & 0  & 0  & 0  \\
DeepSeek-R1-Distill-Qwen-32B  & 53.33              & 100        & 50         & 100        & 100        & 100        & 100 & 25  & 75  & 100 & 0   & 25  & 25  & 0  & 0  & 0  \\
O3-mini (low)                 & 53.33              & 100        & 100        & 100        & 100        & 100        & 100 & 50  & 100 & 25  & 0   & 0   & 25  & 0  & 0  & 0  \\
Gemini-2.0-Flash-Thinking     & 51.67              & 100        & 0          & 100        & 100        & 100        & 100 & 0   & 100 & 50  & 0   & 50  & 50  & 0  & 0  & 0  \\
DeepSeek-R1-Distill-Llama-70B & 50                 & 100        & 50         & 100        & 100        & 75         & 100 & 0   & 50  & 50  & 0   & 25  & 100 & 0  & 0  & 0  \\
DeepSeek-R1-Distill-Qwen-14B  & 50                 & 100        & 50         & 100        & 100        & 100        & 100 & 0   & 75  & 50  & 0   & 0   & 75  & 0  & 0  & 0  \\
QwQ-32B-Preview               & 36.67              & 100        & 25         & 100        & 100        & 0          & 100 & 0   & 100 & 0   & 0   & 0   & 25  & 0  & 0  & 0  \\
Gemini-2.0-Flash              & 30                 & 100        & 0          & 100        & 100        & 0          & 100 & 0   & 50  & 0   & 0   & 0   & 0   & 0  & 0  & 0  \\
DeepSeek-V3                   & 28.33              & 100        & 0          & 50         & 100        & 0          & 75  & 0   & 75  & 25  & 0   & 0   & 0   & 0  & 0  & 0  \\
gemini-2.0-pro                & 26.67              & 100        & 0          & 75         & 100        & 0          & 100 & 0   & 25  & 0   & 0   & 0   & 0   & 0  & 0  & 0  \\
DeepSeek-R1-Distill-Qwen-1.5B & 25                 & 100        & 0          & 100        & 25         & 0          & 100 & 0   & 50  & 0   & 0   & 0   & 0   & 0  & 0  & 0  \\
gpt-4o                        & 13.33              & 100        & 0          & 25         & 0          & 0          & 0   & 0   & 0   & 75  & 0   & 0   & 0   & 0  & 0  & 0  \\
claude-3.5-sonnet             & 3.33               & 25         & 0          & 0          & 0          & 0          & 0   & 0   & 0   & 25  & 0   & 0   & 0   & 0  & 0  & 0  \\
Control-R-32B~(Ours)& 55                 & 100        & 75& 100        & 100        & 100& 100 & 0   & 75  & 100& 0   & 0   & 75  & 0  & 0  & 0 \\
\bottomrule
\end{tabular}
}
\caption{Detailed experiments on AIME2025 Part I Benchmarks, metric is Pass@1(n=4). Reported results cited from MathArena Project~\cite{noauthor_eth-srimatharena_2025}.}
\label{tab:aime2025}
\end{table*}

\end{document}